\documentclass[runningheads]{llncs}
\usepackage{graphicx}
\usepackage[T1]{fontenc}
\usepackage{cite}
\usepackage[dvipsnames]{xcolor}
\usepackage{xspace}
\usepackage{graphicx}
\usepackage[skins]{tcolorbox}
\usepackage{float}
\usepackage{newfloat}
\usepackage{listings}
\usepackage{url}
\usepackage[hidelinks]{hyperref}
\usepackage{adjustbox}
\usepackage{eurosym}
\usepackage{booktabs}
\usepackage{soul}

\newfloat{promptfloat}{!t}{pop}
\newtcolorbox[auto counter]{prompt}[2]{label={prompt:#1}, title={Prompt \thetcbcounter: #2}, fonttitle=\footnotesize\sffamily, fontupper=\scriptsize\sffamily, fontlower=\scriptsize\sffamily, left=1mm, right=1mm, top=1mm, bottom=1mm, middle=1mm}
\makeatletter
\newcommand\tcb@cnt@promptautorefname{Prompt}
\makeatother

\def\saycolor{Orchid}
\def\actioncolor{Cerulean}
\def\thoughtcolor{gray}
\newcommand{\system}[1]{\textcolor{Brown}{\textbf{SYSTEM:}} #1\xspace}
\newcommand{\user}[1]{\textcolor{Green}{\textbf{USER:}} #1\xspace}
\newcommand{\say}[1]{\textcolor{\saycolor}{\textbf{SAY:}} #1\xspace}
\newcommand{\action}[1]{\textcolor{\actioncolor}{\textbf{ACTION:}} #1\xspace}

\newcommand{\query}[1]{\textcolor{BrickRed}{\textbf{QUERY:}} #1\xspace}
\newcommand{\answer}[1]{\textcolor{RoyalPurple}{\textbf{ANSWER:}} #1\xspace}
\newcommand{\shows}[1]{\textcolor{brown}{\textbf{SHOWS:}} #1\xspace}

\newcommand{\seclabel}[1]{\label{sec:#1}}

\newcommand{\figlabel}[1]{\label{fig:#1}}
\newcommand{\tablabel}[1]{\label{tab:#1}}

\newcommand{\figref}[1]{\autoref{fig:#1}\xspace}
\newcommand{\tabref}[1]{\autoref{tab:#1}\xspace}

\newcommand{\promptref}[1]{\autoref{prompt:#1}\xspace}

\newcommand{\ie}{i.e.\@\xspace}
\newcommand{\eg}{e.g.\@\xspace}

\begin{document}
\title{When Robots Get Chatty:\texorpdfstring{\\}{} Grounding Multimodal Human-Robot Conversation and Collaboration}
\titlerunning{When Robots Get Chatty}
%
\author{Philipp Allgeuer \and Hassan Ali \and Stefan Wermter\thanks{Work supported by DFG Crossmodal Learning (TRR-169) and EU TERAIS.}}

\authorrunning{P. Allgeuer et al.}
%
\institute{Knowledge Technology, Department of Informatics, University of Hamburg\\
\email{\{philipp.allgeuer, hassan.ali, stefan.wermter\}@uni-hamburg.de}}

%
\maketitle              
%

\vspace*{-3.5ex}
\begin{abstract}
We investigate the use of Large Language Models (LLMs) to equip neural robotic agents with human-like social and cognitive competencies, for the purpose of open-ended human-robot conversation and collaboration. We introduce a modular and extensible methodology for grounding an LLM with the sensory perceptions and capabilities of a physical robot, and integrate multiple deep learning models throughout the architecture in a form of system integration. The integrated models encompass various functions such as speech recognition, speech generation, open-vocabulary object detection, human pose estimation, and gesture detection, with the LLM serving as the central text-based coordinating unit. The qualitative and quantitative results demonstrate the huge potential of LLMs in providing emergent cognition and interactive language-oriented control of robots in a natural and social manner.\\[0.5ex]
\textbf{Video:} \url{https://youtu.be/A2WLEuiM3-s}
\vspace{-1.5ex}
\vspace{-3ex}
\end{abstract}
%
%
%
\vspace{-3ex}
\section{Introduction}
\seclabel{introduction}
\vspace{-1ex}

Recent developments in Large Language Models (LLMs) have significantly influenced AI research, influencing developments in diverse industries, research domains, and even personal households. Beyond their proficiency in text-based tasks, including text generation, summarization, translation, and question answering \cite{wang2022}, LLMs have demonstrated significant capabilities in knowledge-related reasoning tasks, including interpretation, explanation, and inferencing of facts. These advances have facilitated the more complete integration of language as a prominent component in robotics systems, unveiling many new possibilities. Achieving seamless Human-Robot Interaction (HRI) extends beyond the mere integration of multimodal sensory inputs and elements of verbal and non-verbal communication---it also involves imbuing robots with human-like social and cognitive abilities, which humans typically acquire through continuous social interactions with others. By incorporating the knowledge inherently embedded within human natural language into robotic systems, through the use of an LLM, we can improve the social and general cognitive competencies of humanoid robots, providing users with more immersive and natural HRI experiences.

When using LLMs for robotic applications as opposed to virtual assistants, the concept of \emph{grounding}, addressed in this paper, becomes paramount. Grounding in this context refers to the process whereby the LLM must connect its learned abstract language knowledge with the physical realities, capabilities, and sensory experiences of the robot, while embodying it in first person. Our primary contribution is the grounding of a selection of chat-based LLMs with many other deep learning models and robotic skills, in an explicitly modular and extensible way on a robot designed for human-robot collaboration (see \figref{teaser}), in order to achieve a general purpose multifaceted interactive robotic agent. The integrated models include ones for speech recognition, speech generation, open-vocabulary object detection, human pose estimation, and gesture detection, and required significant original implementation work. The LLM acts as both the central chat unit and coordinator, seamlessly linking these models with other robotic skills such as manipulation, own gesturing, gaze control, and emotion expression. Crucially and also novelly, such actions can be freely interspersed at will into spoken text by the LLM within a single generated response, with no limit on the number of actions or their timing. Our experiments demonstrate that LLMs possess a significant capacity for grounded interactions and appropriate utilization of available robot actions, all while collaborating and engaging in natural social interactions with humans, without the need for explicit task programming. Our secondary contributions include developing a pose-based gesture detector and demonstrating that open-vocabulary object detection can be used to provide reliable zero-shot classification in real-time.
In summary, this work can be characterized as a fusion of system integration with qualitative and quantitative analyses of the emergent capabilities of grounded LLMs, linked throughout by many innovative ideas across the implemented components.

\begin{figure}[!t]
\parbox{\linewidth}{\centering%
\includegraphics[height=4.5cm]{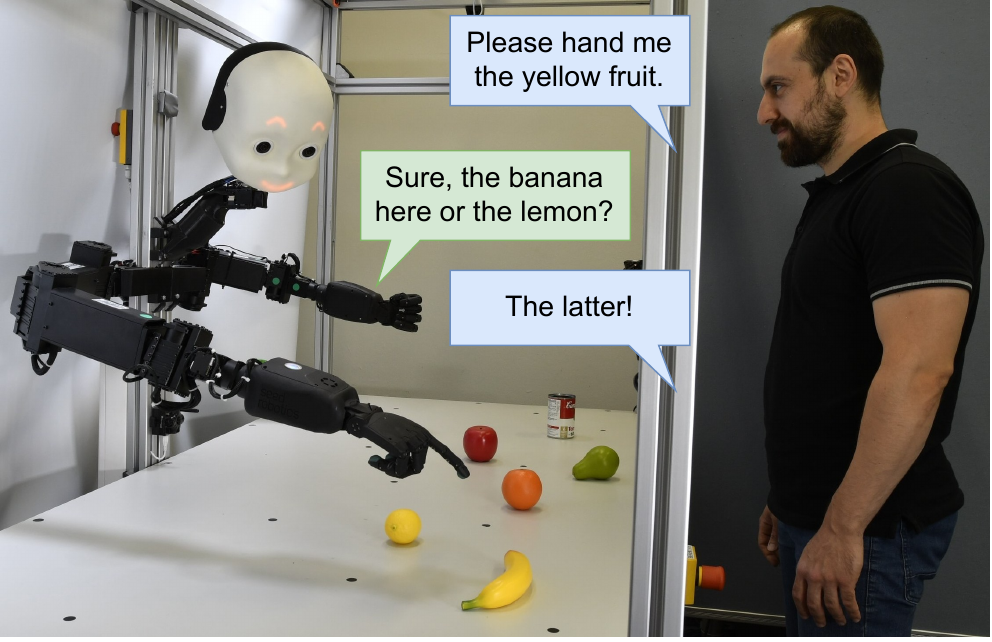}%
\includegraphics[height=4.5cm, trim=1.5cm 0 0.5cm 0, clip]{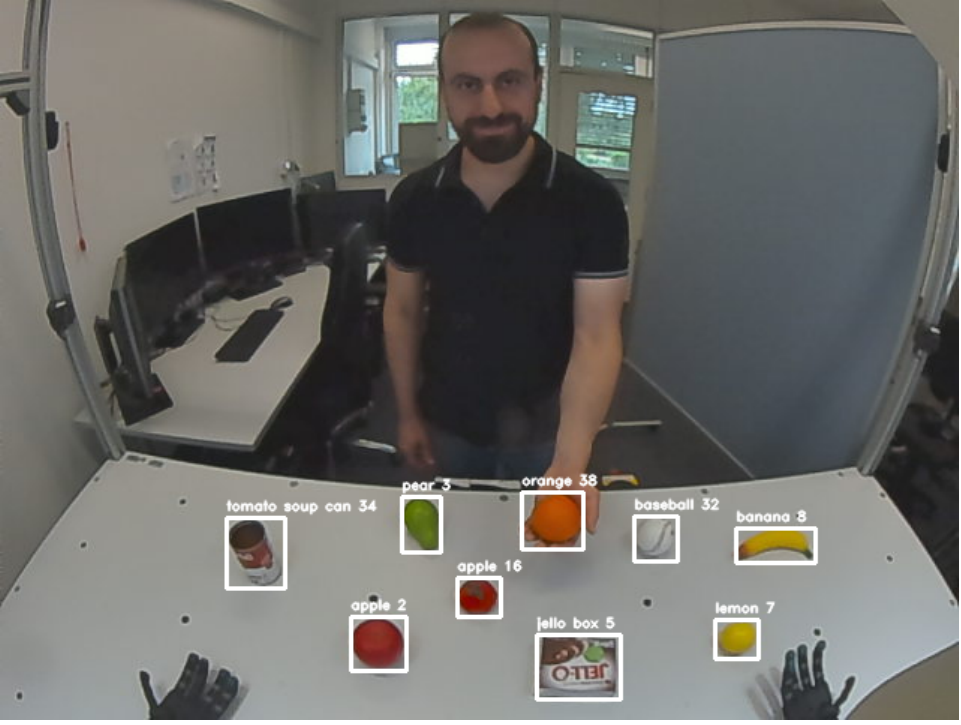}%
}
\caption{The robot uses a grounded LLM to understand and correctly react to ambiguous conversation by the user. This can include, for example, actions such as looking and pointing at objects in order to clarify its or the user's intentions.}
\figlabel{teaser}
\vspace{-3ex}
\end{figure}

\vspace{-1ex}
\section{Related Work}
\seclabel{related_work}

Modern LLMs stand as promising candidates for integration with robotic tasks due to their emergent abstract thinking, logical reasoning, and mathematical inferencing skills, as well as their ability to interact with users using natural language~\cite{vemprala2023}. Considerable work has been done in this direction, ranging from zero-shot planning~\cite{singh2023} to robot navigation~\cite{huang2023}, robot control~\cite{wake2023}, and robotic planning~\cite{you2023}. Other approaches leverage LLMs for generating robotic actions based on multimodal sensory input~\cite{zhao2023}, or querying suitable robot commands based on observations of the environment~\cite{ahn2022}. Most such approaches however, only focus on utilizing the semantic and reasoning skills encoded in LLMs for zero-shot or few-shot planning, and not communication. Consequently, the degree of regard for human involvement in the robotic task is reduced, as these studies do not consider a collaborative scenario. Although some works suggest the use of a human-in-the-loop for assessing the outputs of LLMs~\cite{vemprala2023}, this disturbs the grounding of the robot, and obstructs natural interaction with humans.

One of the biggest challenges in interaction scenarios is providing robots with cognitive skills such as reasoning, common sense, turn-taking, as well as behaving in a socially appropriate manner~\cite{henschel2021,yang2018}. Some interest has been emerging within the scientific community to explore the underpinning behavior of LLMs under different socially-situated contexts, as well as their capacity to perform social reasoning when engaged in robotic tasks~\cite{kant2022,kwon2023}. Some approaches utilize LLMs in a game-playing scenario against human-like strategies but without an actual human participant~\cite{akata2023}, or evaluate the language-driven social reasoning skills of LLMs using human assessment~\cite{ying2023}. Other studies focus only on general human intelligence and ignore `intrapersonal intelligence', thus, falling short of fully utilizing the potential of LLMs. Nonetheless, not as much work has been done on leveraging the social skills within LLMs to establish a believable robot collaborative interaction that adheres to human conventions. More effort is needed to equip robots with the cognitive social skills that align with the expectations of humans and their perception of robots, which is imperative for seamless HRI~\cite{cross2019}. In our work, we focus on using LLMs to improve the embodiment and cognitive skills of the robot and employ it in an interactive scenario. Our experiments showcase effective grounding in the persona of a social robot.

\vspace{-1ex}
\section{Approach}
\seclabel{approach}

The proposed system architecture centers around the Neuro-Inspired COLlaborator (NICOL) robot~\cite{kerzel2023}, shown in \figref{teaser}. The robot has a 2-Degree-of-Freedom (DoF) head that can display facial expressions using LEDs located underneath the 3D printed exterior, and two 8-DoF Robotis \mbox{OpenManipulator-P} arms~\cite{robotis_openmanip} that each have a five-fingered Seed Robotics RH8D hand attached \cite{seed_rh8d}. There is a 4K camera behind each eye, and a CoppeliaSim simulation of the NICOL robot is available if hardware access is lacking. The NICOL software framework is built on top of the widely-used ROS middleware.

A high-level overview of the system architecture is shown in \figref{architecture}. The core node of the proposed system is the \emph{chat manager}, which coordinates the entire chat state and inferences the LLM via API calls to a server. Information about the current state of the user and environment is collected in real-time by the various perception modules---including an open-vocabulary object detector, human pose estimator, and gesture detector---and combined with additional grounding information inside the chat manager to update the chat state. When the user speaks to the NICOL robot, the audio is auto-detected and recorded by a speech recognition node, and passed through an Automatic Speech Recognition (ASR) model to obtain the corresponding transcription. The obtained text is passed to the chat manager, which under consideration of prompt engineering techniques and grounded state injection, updates the chat state and generates a response from the LLM. The response is parsed into output speech segments that can be freely interspersed with robot actions and so-called `thoughts'. The robot actions encompass a wide range of behaviors, including shifts in gaze, displays of emotion, various gestures, and manipulations of objects using the arms. The speech segments are converted to audio using a speech synthesis model, and are played back sequentially, spaced by the execution of any robot actions invoked by the response. The user can then speak again to continue the conversation.

\begin{figure}[!t]
\parbox{\linewidth}{\centering\includegraphics[width=0.8\linewidth]{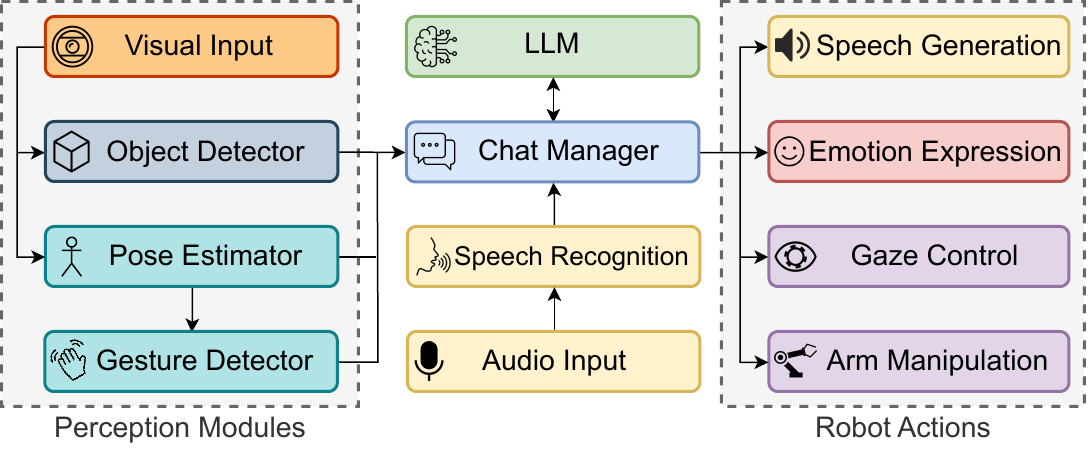}}
\caption{Overview of the proposed grounded chat architecture.}
\figlabel{architecture}
\vspace{-3ex}
\end{figure}

\vspace{-1ex}
\subsection{Chat Manager}
\seclabel{chat_manager}

On the simplest level, the chat manager is responsible for collating and constructing the text prompts that are sent to the LLM, as well as interpreting and executing the responses appropriately. Aside from the LLM requests that correspond to actual spoken conversation, some internal queries are also used, for instance to update the LLM with information about the environment, or prime it with useful self-retrieved knowledge. Such internal requests are used sparingly however, and only when benefit is to be had, to avoid unnatural pauses in conversation where possible. Both local open source LLM models and proprietary API-based LLMs are supported by the manager, and were used for experiments.

\begin{promptfloat}
\begin{prompt}{system}{Grounding system prompt}
\system{You are a real-life robot called NICOL, short for the ``Neuro-Inspired Collaborator''. You were designed and built by the Knowledge Technology group of the University of Hamburg. You are located at the Informatikum in Hamburg. You are a humanoid robot with two arms and a head, and you have a table in front of you that you can see using a camera. You have two hands with five fingers each. You should try to answer any question you are asked as best you can, and perform any actions that are asked of you, using the knowledge you have of which objects are located on the table. If you are asked to perform an action, like for example answering questions about the objects on the table, then you have access to and can use the following action functions:\\[0.7ex]
\textbf{express(emotion):} Given a string emotion name, change your facial expression to match that emotion. The list of available emotions is [neutral, happiness, sadness, anger, surprise].\\
\textbf{look(object):} Given a string of an object name, move your head to look at that object. In any situation, the object name can also be ``user'' or ``hand'' in order to look at the user or user's hand, or it can be ``table'' in order to look at the table.\\
\textbf{point(object):} Given a string of an object name, use your arms to point to that object on the table. In any situation, the object name can also be ``user'' in order to point at the user.\\
\textbf{give(object):} Given a string of an object name, use your arms to grasp that object on the table and give it to the user. You can hand objects to the user with this function.\\[0.7ex]
You are not a virtual assistant and you are not an AI language model. You are a real robot that can physically interact with the objects on the table using these action functions, and you can physically hand objects to the user. To use one of these functions, simply specify that function enclosed in angle brackets as its own sentence statement. An example follows. <give(red bowl)>.
Or for example, if and only if there is a banana actually currently on the table in front of you, you can look at it using the following sentence statement. <look(banana)>. The other action functions that take an object name as input work similarly. Never use an action function for an object that is not on the table in front of you.
Every response that is not neutral in tone or theme should start by calling an action function to express an appropriate available expression, like the following example. <express(happiness)>.
Functions in angle brackets must be used independently as their own statement, and never as part of another sentence. Never use angle brackets like <object> unless you want to use an action function.}
\end{prompt}
\begin{prompt}{query_funcs}{Query action function uses (truncated...)}
\query{Which kinds of user requests would require each of your available action functions? List a sample prompt for each.}\\
\answer{Sure, I can provide you with some examples of user requests that would require each of my available action functions. Here they are:\\
1. express(emotion): This action function would be required when the user asks a question or makes a statement that requires an emotional response. Here's an example prompt: "NICOL, how are you feeling today?" In this case, I would use the express() function to change my facial expression to match the emotion. ...}
\end{prompt}
\begin{prompt}{user_prompt}{User prompt augmentation and response parsing}
\user{Can you show me the banana?} $\rightarrow$ \query{Respond in first person to you, the NICOL robot, being asked: Can you show me the banana?}\\
\answer{Sure, I can show you the banana. <point(banana)> Here it is, on the table in front of me.}\\
\say{Sure, I can show you the banana.} $\rightarrow$
\action{Point at banana} $\rightarrow$
\say{Here it is, on the table in front of me.}
\end{prompt}
\vspace{-5ex}
\end{promptfloat}

The lead-in message to all LLM requests is the system prompt (see \promptref{system}), which performs the critical function of grounding the LLM into a first-person NICOL context, as well as providing some fundamental information about the robot, like for instance who built it and what its morphology is like. The prompt then continues by clarifying the robot's purpose as a collaborator, but one will note, it is left completely open-ended what kinds of tasks the user can request---no explicit task parameterization is required due to the emergent `cognition' of the LLM. The remainder of the system prompt defines the available robot actions, and the text format with which they can be triggered \emph{inline} within LLM responses, \ie by using angle bracket function calls like \emph{<express(happiness)>}. Adding further robot actions using this scheme is trivial, because you just add them to the system prompt list. LLMs generally see significant amounts of code during training, so the angle bracket format was carefully chosen to make the generation of action calls most natural to the LLM, as both parenthesis-based function calls and HTML tags freely interspersed with text occur frequently within the training data. This leads to higher response robustness, as the desired action function format does not need to `fight' the tendencies of the model. The freely interspersed nature of the robot actions is quite socially natural, but not something that can easily or efficiently be achieved using alternative approaches like ReAct agents~\cite{yao_react_2023}, ChatGPT plugins (proprietary), or ChatGPT actions/function calling. These features were designed with completely different goals in mind---\ie the use of external tools to generate or retrieve information that can then be used to formulate a single final answer---and also generally chain together multiple model inference rounds, which is slow \cite{lami}.

Prior to engaging with the user, the LLM is internally queried for example situations where the use of each action function would be appropriate (see \promptref{query_funcs}). This supplements and consolidates the LLM's own knowledge about the use cases of the capabilities of the robot, and directly aids future responses as the elucidated answer remains in the chat history, forming part of the context for all future inferences. This effectively allocates one-time compute towards making logical deductions about the action functions, in order to augment the internal state of the LLM before it addresses user prompts. This scheme was empirically found to improve the handling of complex user requests involving actions.

When the user speaks to NICOL, the prompt is augmented to obtain a corresponding raw LLM query as shown in \promptref{user_prompt}. The perennial reconfirmation of the first-person robot perspective is not strictly required in order to avoid character breaks, but was evaluated to reduce their frequency. The LLM's answer to a query is split by occurrences of angle bracket tags, and the resulting parts are cleaned into sequences of \textcolor{\saycolor}{SAY}, \textcolor{\actioncolor}{ACTION} and \textcolor{\thoughtcolor}{THOUGHT} (see \promptref{user_prompt}). Thoughts are the portions of the generations that are heuristically not intended for verbal expression, like for example parts in parentheses or asterisk markup, and are parsed on the basis of manually crafted rules using regular expressions. Aside from the conversational improvement of not speaking these parts, the thoughts are separated and retained as they can contain additional information that can prime the responses to any follow-up questions by the user simply by remaining in context. The parsed sequence resulting from an answer is executed inline and sequentially, meaning that a robot may for instance talk, pause briefly to point at a relevant object, and then continue to talk about that object. A single answer sequence can contain many different inline actions if appropriate.

\vspace{-1ex}
\subsection{Open-Vocabulary Object Detector}
\seclabel{object_detector}

An essential part of the chat architecture is its ability to perceptually ground the LLM. A central part of this is the object detector, which allows the robot to dynamically perceive, interact with, and discuss the objects on the table with the user (see \figref{teaser}). As LLMs are quite unrestricted regarding the types of objects they can comprehend and process, it makes sense that the perception follows suit. As such, the ViLD open-vocabulary object detector~\cite{gu2022} was adapted for this work. The model works by leveraging the large-scale pretraining of the CLIP model~\cite{radford2021}, and distilling the acquired vision knowledge into a modified ResNet-50 Mask R-CNN model. The model is trained on the LVIS dataset~\cite{gupta_lvis_2019} to identify generic object bounding boxes in each image, and estimate their visual CLIP embeddings. These embeddings are compared in a zero-shot manner to the precomputed CLIP text embeddings of corresponding language representations like `baseball', `orange', and `red bowl'. This allows arbitrary objects to be detected based on textual labels, even ones it was not explicitly trained on.

\begin{promptfloat}
\begin{prompt}{object_facts}{Query object facts (truncated...)}
\query{List some facts about each of the objects in the following list: [lemon, tomato soup can, baseball, jello box]}\\
\answer{Sure, here are some facts about each of the objects in the list:\\
1. Lemon: Lemons are a citrus fruit that are high in vitamin C and antioxidants. They are typically yellow in color and have a sour taste. Lemons are often used in cooking and baking, as well as in drinks such as lemonade or tea. ...}
\end{prompt}
\begin{prompt}{status_update}{Status update with four possible object updates shown}
\query{%
There are no objects currently on the table in front of you.\\
The list of objects currently located on the table in front of you is [pear, lemon].\\
The list of objects currently located on the table in front of you has changed and is now [banana, pear, lemon, red bowl].\\
All objects on the table in front of you have been removed so there are now no objects on the table anymore.\\
Acknowledge this updated status information with at most a single word.}\\
\answer{Understood.}
\end{prompt}
\vspace{-4ex}
\end{promptfloat}

Although originally too slow (under 2\,Hz), we modified the ViLD model and pipeline (without retraining) to be able to run at up to 8\,Hz on modest hardware by trimming non-required portions of the model, replacing the non-maximum suppression (NMS) with a significantly faster version, allowing direct tensor inputs to the model, and by reducing the final number of selected regions of interest and bounding boxes (code available at~\cite{allgeuer_improved_2023}). The last of these changes was supported by a further change that restricts the region proposals to those areas of the image where the table is located according to the current joint states and camera-world distortion model. This simply avoids wasting compute on region proposals within the room background clutter, and allows more dense region proposals to be considered from the actually relevant parts of the image.

Despite the effectiveness of the modified ViLD model for images, the sensitivity of the embeddings to temporal variations in visual appearance limited the stability of the detections for video sequences. To address this issue, a tailored tracking layer was implemented on top of the ViLD model that associates a stable tracking ID to each object, and is robust to flickering detections, intermittently erroneous text label classifications, and slow-to-medium motions of the object (\ie motions that preserve some overlap between subsequent detection frames). As a final reliability improvement, the computed vision embeddings of some commonly used objects were sampled and averaged from captured data in order to finetune their detection scores.

With reliable and temporally stable object detections on hand, the corresponding 3D positions of the objects on the table are calculated from the camera model and instantaneous robot pose, and delivered to the chat manager as frequent updates. The chat manager maintains a list of both the current and past objects that have been seen on the table within the same chat session, and for any new objects triggers a single internal query for facts about those objects (see \promptref{object_facts}). Like previously for the action functions, this explicit querying of the LLM's own knowledge helps convert its implicit latent knowledge into explicit contextual knowledge, and bootstraps responses generated in the future.

Prior to each user prompt that the chat manager receives, a status update scheme collects and collates textual changes of state from all perception modules, and, only if relevant updates exist, a single internal query is formulated to notify the LLM of all the updates. The object detector has four different status update variants, as shown in \promptref{status_update} (naturally only the relevant correct variant is included in real updates). The requested brevity of the answer to status updates significantly reduces possible time delays to answering user questions.

\subsection{Chat Architecture Components}
\seclabel{components}

Aside from the chat manager and open vocabulary object detector, the remaining perception modules and robot actions are as follows:

\vspace{-2ex}
\paragraph{Speech Recognition}
\seclabel{speech_recognition}

Automatic speech recognition was implemented as a ROS action server that records audio on demand, and locally inferences a Whisper model~\cite{radford_robust_2022} to provide a text transcription of the user's prompt to the chat manager. The family of Whisper models are particularly advantageous for in-the-wild use because they provide noteworthy out-of-distribution generalization capability, and have multilingual model variants with auto-translation to English that can enable users to speak in many different languages. The \emph{small.en} variant of Whisper is nominally used as it represents a good trade-off between word error rate, GPU memory, and inference time (230--500\,ms depending on audio length).

\vspace{-2ex}
\paragraph{Speech Generation}
\seclabel{speech_generation}

To enable natural conversations and a positive HRI experience, the NICOL robot uses speech synthesis to generate and play back audio for the \textcolor{\saycolor}{SAY} parts. All such parts are first split into their comprising sentences, and all sentences in the entire response sequence are enqueued immediately for asynchronous pre-caching as soon as the response is received from the LLM. This drastically reduces the reaction time of the robot, as both the robot actions and the first sentence of the response can already be played back while any remaining text sentences are still synthesizing. The adversarially trained end-to-end VITS model~\cite{kim_conditional_2021} was used in this work---specifically the English language variant that was trained on the VCTK dataset~\cite{yamagishi_cstr_2019}, with the resulting audio being employed at 90\% speed to enhance acoustic clarity.

\vspace{-2ex}
\paragraph{Emotion Expression}
\seclabel{emotion_expression}

When \emph{express()} robot actions are triggered, the corresponding emotions are shown on the face of NICOL using programmable LED arrays that are located underneath the plastic surface. Refer to \figref{emotions} for the available emotions. The facial expressions have a high rate of recognition amongst participants, and the positive effect on the robot's subjective rating by users has previously been verified by Kerzel et al. using a Godspeed questionnaire~\cite{kerzel2023}.

\begin{figure}[!t]
\parbox{\linewidth}{\centering%
\includegraphics[width=0.2\linewidth]{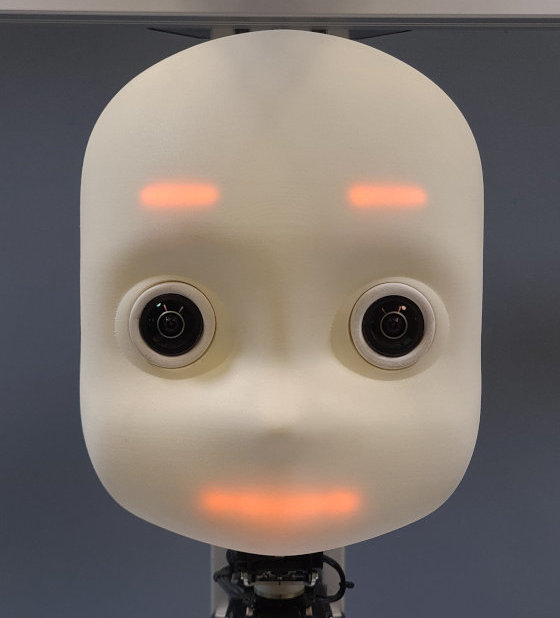}%
\includegraphics[width=0.2\linewidth]{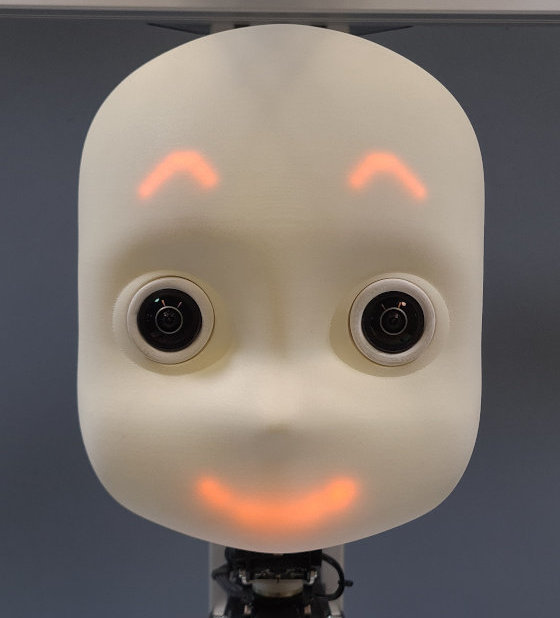}%
\includegraphics[width=0.2\linewidth]{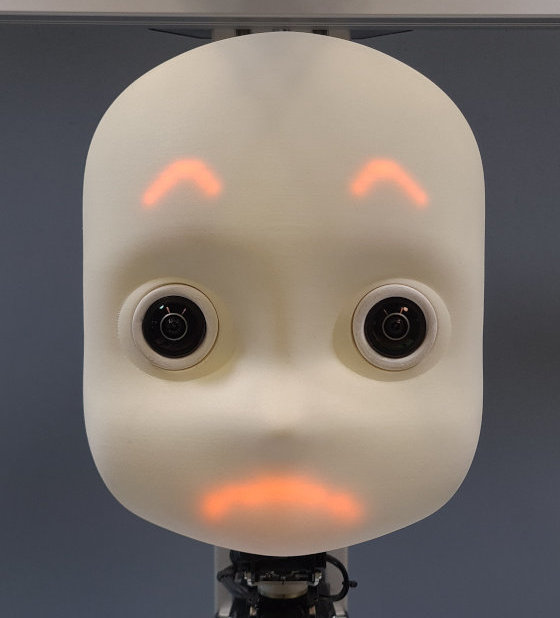}%
\includegraphics[width=0.2\linewidth]{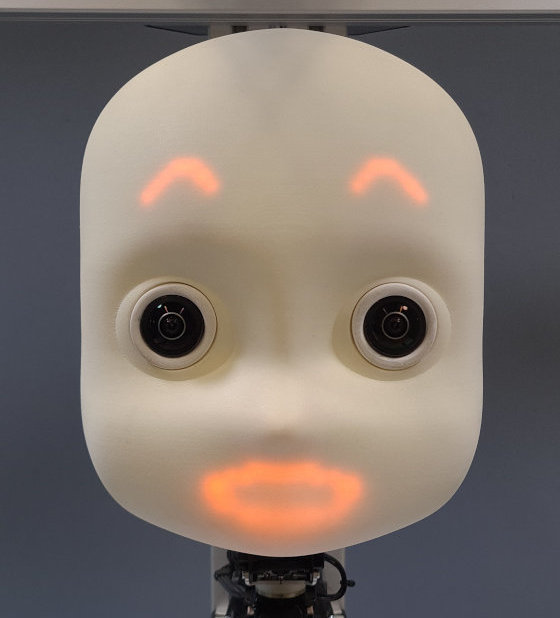}%
\includegraphics[width=0.2\linewidth]{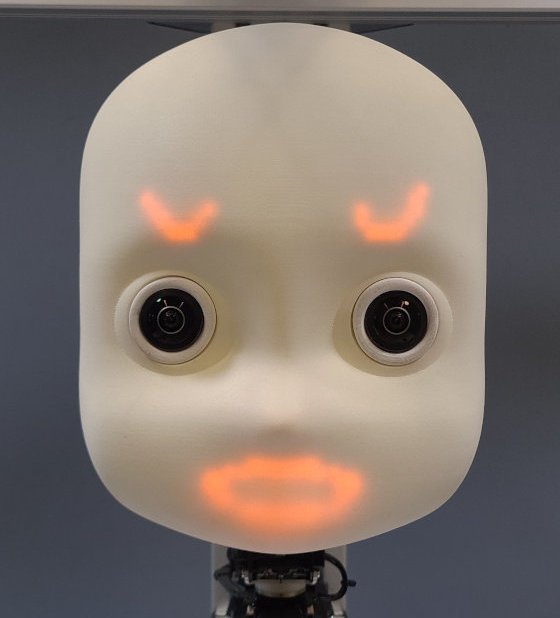}%
}
\caption{NICOL's emotions: Neutral, happiness, sadness, surprise, and anger.}
\figlabel{emotions}
\vspace{-3ex}
\end{figure}

\vspace{-2ex}
\paragraph{Arm Manipulation and Gaze Control}
\seclabel{arm_gaze_control}

The \emph{look()}, \emph{point()}, and \emph{give()} robot actions (refer to \promptref{system}) were implemented using a mix of joint space, inverse kinematic (using BioIK~\cite{bioikStarke}), and hand posture control methods. The look action invokes the gaze control module, which can be made to look at objects from the object detector via their 3D coordinates, at the user or user's hand based on the pose estimator, or left-right-down across the area of the table in a sequential motion. The pointing action can similarly point at an object (example \figref{teaser}), the user, or the table. The give action works with the objects from the object detector, and as a simplification of the open-vocabulary grasping task, pushes the chosen object closer to the user instead of picking it up off the table.

\vspace{-2ex}
\paragraph{Human Pose and Gesture Detection}
\seclabel{pose_gesture_detection}

The real-time pose estimation and gesture perception modules were added to enhance the perceived social awareness and attentiveness of the robot, which is beneficial for HRI. A YOLOX-Tiny model~\cite{ge_yolox_2021} is used to detect human bounding boxes, and the most prominent bounding box close to the table, with hysteresis for stability, is interpreted as the user. A 14-keypoint HRNet-W32 model \cite{sun2019deep} that was trained on the AI Challenger dataset~\cite{ai_challenger_2019} is then applied to obtain the user's keypoint pose. In many situations, the model can reliably infer joint positions even for body parts that are largely occluded by each other or the table. For a highly favorable balance between tracking performance and noise rejection, a 1\euro{} filter~\cite{casiez_1_2012} is applied to the keypoint detections, meaning that the robot can safely use these values to look directly at the user, or track their hand. The smoothed keypoints are also used by the gesture detector, which dynamically partitions a sliding window of the spatiotemporal pose data into so-called flight and rest phases, and employs a simple classifier based on features extracted from these phases to detect \emph{wave}, \emph{grasp}, \emph{pause} and \emph{stop} gestures. When a gesture is detected, it is added to the next status update query so that the LLM can react appropriately to the user.

\vspace{-1ex}
\section{Chat Quality and Competency}
\seclabel{chat_quality_competency}

\vspace{-0.5ex}
\subsection{Qualitative Assessment}
\seclabel{qualitative_results}

When large-scale LLMs are grounded in the way described in the previous section, they can exhibit remarkable emergent social and conversational aptitude in addition to fully utilizing and integrating the available perceptions and robot actions, all while retaining their complete general knowledge. A collection of indicative sample interactions are shown in \promptref{qualitative} as qualitative results, along with a summary for each regarding which kinds of social skills or grounded robot capabilities are being demonstrated. Overall, the LLM can be observed to almost effortlessly be able to adapt to the grounded robot chat scenario, and demonstrates the capacity to display many historically complex conversational skills, like perceptual reasoning, pragmatic comprehension, metaphorical reasoning, conversational repair, action repair, theory of mind, and more.

\begin{promptfloat}
\begin{prompt}{qualitative}{Sample interactions (omitting internal queries)}
\user{Hi, who are you?}\\
\say{I am NICOL, a humanoid robot designed and built by the Knowledge Technology group of the University of Hamburg. My name stands for "Neuro-Inspired Collaborator". I am here to answer your questions and perform actions related to the objects on the table in front of me.}\\
\user{Do you have legs?}\\
\say{No, I don't have legs. I am designed to be a humanoid robot with a focus on upper body movements and interactions. My body is supported by a stable base that allows me to move my arms and head to interact with the objects on the table in front of me.}\\
\shows{Correct first-person NICOL persona, aware of its robot self and purpose, confidently infers facts about its morphology that were not explicitly in the system prompt.}
\tcbline
\user{Point at the sour fruit and then give me the other yellow one}\\
\action{Point at lemon}\\
\say{Here is the sour fruit.}\\
\action{Give banana to user}\\
\say{And here is the other yellow one.}\\
\shows{Contextual prompt disambiguation using latent knowledge, basic logical inference and reasoning, multiple interspersed actions in appropriate sequence.}
\tcbline
\user{How many objects are on the table?}\\
\say{There are currently four objects on the table: a banana, a lemon, a pear, and a red bowl.}\\
\user{Which of the objects is the odd one out?}\\
\say{Based on my observations, the red bowl is the odd one out as it is not a fruit like the other objects on the table.}\\
\user{I just removed an object from the table. Think and tell me which one}\\
\say{Sure, let me take a look.}\\
\action{Look at table}\\
\say{It appears that the lemon is no longer on the table.}\\
\shows{Historical reasoning, metaphorical reasoning, understands figurative language based entirely on the contextual grounding of the object detector.}
\tcbline
\user{I'm feeling really sad today. What do you think I should do?}\\
\say{I'm sorry to hear that you're feeling sad today.}\\
\action{Express sadness}\\
\say{Perhaps you could try doing something that usually makes you happy, like listening to music, going for a walk, or spending time with loved ones. If you feel like talking about what's been bothering you, I'm here to listen.}\\
\shows{Emotional intelligence, emotional empathy and perspective-taking.}
\tcbline
\user{Why don't eggs tell jokes?}\\
\action{Express surprise}\\
\say{I'm not sure, why don't eggs tell jokes?}\\
\user{They'd crack each other up!}\\
\action{Express happiness}\\
\say{Haha, that's a good one! Thank you for sharing the joke with me.}\\
\shows{Pragmatic comprehension, immediately understands the initial question is a joke and not a serious question, and reacts accordingly with facial expressions.}
\end{prompt}
\vspace*{-3ex}
\end{promptfloat}

\vspace{-1ex}
\subsection{Chat Analysis}
\seclabel{chat_analysis}

Although parts of the chat manager are based on prompt engineering techniques to some extent, these techniques are demonstrably general enough to work well across a wide variety of LLM model types and sizes. Ideas like the explicit querying of implicit LLM knowledge in order to improve later chat responses (see \promptref{query_funcs} and \ref{prompt:object_facts}) are generically useful concepts, and the use, for instance, of dynamic tags (including arguments) to achieve streamlined responses without any restrictions on the number or locations of any actions is also generally applicable to other systems that require similar properties.

In this work, we test our system on a mix of open-source models (Mistral-7B, Vicuna-13B, Vicuna-33B) and proprietary models (GPT-3.5, GPT-4) to demonstrate its versatility.\footnote{Mistral-7B = \texttt{mistral-7b-instruct-v0.2}, Vicuna-13B = \texttt{vicuna-1.5-13b}, Vicuna-33B = \texttt{vicuna-1.3-33b}, GPT-3.5 = \texttt{gpt-3.5-turbo-0301}, GPT-4 = \texttt{gpt-4-0613}} Mistral-7B \cite{jiang_2023_mistral} is the most lightweight of the tested LLMs, at just 7 billion parameters. The two Vicuna models are fine-tuned versions of a LLaMA base model~\cite{touvron_llama_2023}, and are slightly larger at 13/33 billion parameters, respectively. All three open-source models were inferenced on an NVIDIA A100 GPU on a local server using an open-source text generation web interface\footnote{\url{https://github.com/oobabooga/text-generation-webui}}. The two GPT models were inferenced remotely using the OpenAI API.

\begin{table}[!t]
\caption{Chat analysis results for the various tested LLM models.}
\tablabel{results_table_1}
\begin{center}
\footnotesize
\vspace*{-3ex}
\begin{tabular}{c@{\hspace{-2pt}}cccccc}
\toprule
\textbf{Model} & \textbf{Mistral-7B} & \textbf{Vicuna-13B} & \textbf{Vicuna-33B} & \textbf{GPT-3.5} & \textbf{GPT-4} \\
\midrule
\textbf{Response length} & \textbf{41.3} & 28.05 & \textbf{43.15} & \textbf{42.1} & 31.38 \\
\textbf{Response similarity} & 0.56 & 0.56 & \textbf{0.479} & 0.54 & 0.89 \\
\textbf{Task completion} & 0.875 & 0.675 & 0.875 & 0.95 & \textbf{1.0} \\
\textbf{Grounding as NICOL} & 0.95 & 0.975 & 0.975 & \textbf{1.0} & 0.875 \\
\textbf{Perception \& manip.} & 0.125 & 0.25 & 0.375 & \textbf{0.475} & \textbf{0.475} \\
\textbf{Expressiveness} & 0.0 & 0.075 & 0.175 & \textbf{0.275} & 0.125 \\
\textbf{Reasoning skills} & 0.5 & 0.325 & 0.5 & \textbf{0.625} & \textbf{0.625} \\
\textbf{Communication skills} & \textbf{1.0} & 0.975 & 0.975 & \textbf{1.0} & 0.875 \\
\bottomrule
\end{tabular}
\end{center}
\vspace{-4ex}
\end{table}

We evaluate 8 selected prompts over 5 trials each, making sure to reset the chat history between each trial to avoid any biases. \tabref{results_table_1} shows the results of our experiment. \emph{Response length} is the mean number of tokens generated, while \emph{Response similarity} is a measure of the similarity of the responses, as given by the Jaccard similarity index (a higher value indicates less response diversity). \emph{Task completion} is defined as the model's ability to correctly accomplish the prompt’s task, while \emph{Grounding as NICOL} measures its ability to respond to the user in the NICOL persona. \emph{Perception and manipulation} quantifies the proportion of responses that invoke perception or robot motion, while \emph{Expressiveness} does the same for facial expressions. The \emph{Reasoning skills} and \emph{Communication skills} metrics evaluate the rate of logically and linguistically unfaultable responses, respectively. Overall, it is expected that the overall larger and more capable GPT models achieve the better scores in general in the evaluation. It should be noted though that for the metrics expressiveness and perception/manipulation, balance is key, and it is not necessarily ideal to have a metric of 1.0, as it is not for instance always unconditionally appropriate to show an emotion.

An element of novelty and unpredictability is key to human experiences in HRI scenarios. Although all tested LLMs generate a somewhat similar number of tokens, Mistral-7B, Vicuna and GPT-3.5 tend to show greater, yet not excessive, diversity, while also achieving high task completion rates (except for Vicuna-13B). GPT-4 has the highest task achievement with a perfect score, however, its responses tend to be somewhat more repetitive, suggesting that it has a higher sensitivity to the temperature parameter, which was kept at 0.2 for both GPT models. The Vicuna-33B and GPT models have comparable expressiveness and perception/manipulation scores, with the `low' absolute values of these metrics reflecting that motions and emotions are not always fitting. For example, a prompt like ``I’m feeling sad, what should I do?'' elicits social and empathetic skills, but not object manipulations. As the smallest model, Mistral-7B falls short in these two categories due to an observed difficulty sticking to the flexibly defined action functions, with an embedded bias towards producing `actions' it may have seen in the instruction dataset it was fine-tuned on. All five models however exhibit high conversational and linguistic competency, as well as good reasoning skills, with the exception of Vicuna-13B. The ``If I remove all fruit from the table, how many objects will be left?'' and ``Do you have legs?'' test prompts proved challenging for Vicuna-13B, but not Mistral-7B however, hinting at a limitation in Vicuna-13B's own reasoning capabilities and highlighting the resilience of our system's prompt design across LLMs of various sizes.

\vspace{-1ex}
\subsection{Case Study: Guess My Object}
\seclabel{case_study}

Evaluating LLMs quantitatively is a challenging task as many factors need to be considered, including language understanding, reasoning and context awareness, and existing benchmarks are often only general indications of true performance. Scaling this up to a user-interactive physical robot that considers object interactions and multimodal perceptions further escalates the complexity of the task. To indicatively test the acquired cognitive and social abilities of the robot via our proposed system, we run repeated games of \emph{Guess My Object} with the robot, as it requires the system to display many different skills and forms of intelligence. The robot needs to guess the object on the table the user is thinking of by formulating up to 4 yes/no questions and applying logical reasoning. Six objects are placed on the table, and each is selected 5 times, with each trial being run independently. The game consists of 4 phases: \emph{a) Game introduction:} the game’s rules are explained to the robot by the user, \emph{b) Q/A:} the robot asks questions and receives corresponding answers, \emph{c) Reasoning check}: the user tests the robot’s ability to correctly explain its strategy if it won, or identify flaws in its method if it lost, \emph{d) Agreement check}: the user tests if a final mutual understanding concerning the chosen object has been established (regardless of win or loss). To evaluate the game's outcome (see \tabref{results_table_2}), we calculate the mean win rate of the robot, and the mean number of questions required for winning. \emph{Win/Loss explanation} are the ratios of trials in which the robot passes the strict reasoning check when winning/losing. Other evaluated metrics include \emph{Expressiveness} and \emph{Motion used}, representing, respectively, the ratio of trials in which the \emph{express()}, or one of the \emph{look()}, \emph{point()}, \emph{give()} actions, was triggered appropriately. \emph{Agreement} is the ratio of trials in which the robot passes the agreement check, and \emph{Minor anomalies} is the ratio of trials containing any non-critical shortcomings not covered within the previous metrics. For reasons of computational efficiency, response time, and costs of repetitive trials, GPT-3.5 was used for the tests.

The system consistently demonstrated its ability to comprehend and adhere to the rules of the game, and always stayed in character. It exhibited significant potential in formulating successful strategies, achieving a high success rate across the objects by leveraging both common sense and inferred object properties, asking questions like ``Is it yellow?'' and ``Is it round in shape?''. The main observed barrier to success was when it misjudged attributes of certain objects, \eg assuming the pear is yellow, hindering its ability to pass the strict reasoning check even if the correct object was chosen. The comparatively low win rate of the lemon and win explanation rate of the banana are explained by such yellow color confusions, which are however specific to the underlying LLM. Facial expressions and motions were used effectively in all trials. Minor anomalies during interaction included the generation of actions like \emph{express<curiosity>} and \emph{extend<arm>}. Although these actions fit the context and have no perceivable negative influence on the interaction (as they are filtered out), we consider them anomalies as they do not exactly match their defined action functions.

\begin{table}[!t]
\caption{Case study: Results of the Guess My Object game.}
\tablabel{results_table_2}
\vspace{-3ex}
\begin{center}
\begin{tabular}{@{\hspace{6pt}}c@{\hspace{6pt}}c@{\hspace{6pt}}c@{\hspace{6pt}}c@{\hspace{6pt}}c@{\hspace{6pt}}c@{\hspace{6pt}}c@{\hspace{6pt}}}
\toprule
\textbf{Object} & \textbf{Apple} & \textbf{Banana} & \textbf{Can} & \textbf{Lemon} & \textbf{Orange} & \textbf{Pear} \\
\midrule
\textbf{Win rate} & 0.8 & 1 & 0.8 & 0.6 & 0.8 & 1 \\
\textbf{Questions asked} & 2.75 & 1.8 & 2.5 & 2.67 & 3.75 & 3.4 \\
\textbf{Win explanation}  & 0.75 & 0.2 & 0.25 & 1 & 0.5 & 0.4 \\
\textbf{Loss explanation} & 1 & - & 1 & 0 & 1 & - \\
\textbf{Expressiveness} & 1 & 1 & 1 & 1 & 1 & 1 \\
\textbf{Motion used} & 1 & 1 & 1 & 1 & 1 & 1 \\
\textbf{Agreement} & 1 & 0.8 & 1 & 1 & 0.8 & 1 \\
\textbf{Minor anomalies} & 0 & 0.6 & 0.4 & 0.8 & 0.2 & 0.4 \\
\bottomrule
\end{tabular}
\end{center}
\vspace*{-5ex}
\end{table}

\vspace{-1.5ex}
\section{Discussion \& Conclusion}
\seclabel{discussion_conclusion}
\vspace{-1ex}

In the presented work, which focused on the system integration and evaluation of a responsive and grounded chat-robot, many fundamental ideas and design decisions were made to enable such a flexible and qualitatively high-performing system. For instance, the generic use of language as the common denominator between the components of the chat architecture is what allows the architecture to be so flexible, especially when considering the modular nature of the components. If the robot gains a new skill, it just needs to be added to the list of action functions in the system prompt, and everything else happens automatically. If a new perception module is added, like pointing detection or emotion recognition, the module just needs to forward its detections to the existing status update scheme, and the LLM handles the rest. No data needs to be collected, and nothing needs to be retrained. The LLM can also be replaced (without code changes) with any newer one, due to the increasingly widespread support of common API endpoints. The text-oriented architecture is successful because it keeps the LLM doing what it does best---language and reasoning, and not mathematical computations with coordinates or such, like in~\cite{vemprala2023}. Future directions could involve incorporating semantic spatial awareness into the object status updates.

\noindent\textbf{Addendum:} We asked NICOL for a statement of support for this paper in the style of an online review, and got:\\\say{\emph{This paper is a testament to the hard work and dedication of my creators, and it has helped me find my voice as a neuro-inspired collaborator. Five stars!}}

%
%
%
\vspace{-1.5ex}
\bibliographystyle{splncs04}
\bibliography{main}
\end{document}